\documentclass{article}

\usepackage{arxiv}

\usepackage[utf8]{inputenc} 
\usepackage[T1]{fontenc}    
\usepackage{hyperref}       
\usepackage{url}            
\usepackage{booktabs}       
\usepackage{amsfonts}       
\usepackage{nicefrac}       
\usepackage{microtype}      
\usepackage{lipsum}		
\usepackage{graphicx}
\usepackage{natbib}
\usepackage{doi}
\usepackage{amsmath}

\title{Comparing YOLOv11 and YOLOv8 for instance segmentation of occluded and non-occluded immature green fruits in complex orchard environment}

\author{
  \href{https://orcid.org/0000-0002-5417-6744}{Ranjan Sapkota} \\
  Cornell University\\
  \texttt{rs2672@cornell.edu} \\
  \And
  \href{https://orcid.org/0000-0001-5337-4848}{Manoj Karkee} \\
  Cornell University\\
  \texttt{mk2684@cornell.edu} \\
}

\renewcommand{\shorttitle}{YOLOv11 vs YOLOv8 Instance Segmentation }

\hypersetup{
pdftitle={A template for the arxiv style},
pdfsubject={q-bio.NC, q-bio.QM},
pdfauthor={David S.~Hippocampus, Elias D.~Striatum},
pdfkeywords={First keyword, Second keyword, More},
}

\begin{document}
\maketitle

\begin{abstract}
	This study conducted a comprehensive performance evaluation on YOLO11(or YOLOv11) and YOLOv8, the latest in the "You Only Look Once" (YOLO) series, focusing on their instance segmentation capabilities for immature green apples in orchard environments. YOLO11n-seg achieved the highest mask precision across all categories with a notable score of 0.831, highlighting its effectiveness in fruit detection. YOLO11m-seg and YOLO11l-seg excelled in non-occluded and occluded fruitlet segmentation with scores of 0.851 and 0.829, respectively. Additionally, YOLO11x-seg led in mask recall for all categories, achieving a score of 0.815, with YOLO11m-seg performing best for non-occluded immature green fruitlets at 0.858 and YOLOv8x-seg leading the occluded category with 0.800. In terms of mean average precision at a 50\% intersection over union (mAP@50), YOLO11m-seg consistently outperformed, registering the highest scores for both box and mask segmentation, at 0.876 and 0.860 for the "All" class and 0.908 and 0.909 for non-occluded immature fruitlets, respectively. YOLO11l-seg and YOLOv8l-seg shared the top box mAP@50 for occluded immature fruitlets at 0.847, while YOLO11m-seg achieved the highest mask mAP@50 of 0.810. Despite the advancements in YOLO11, YOLOv8n surpassed its counterparts in image processing speed, with an impressive inference speed of 3.3 milliseconds, compared to the fastest YOLO11 series model at 4.8 milliseconds, underscoring its suitability for real-time agricultural applications related to complex green fruit environments. 
\end{abstract}


\section{Introduction}
In the rapidly evolving domain of artificial intelligence (AI), advancements in object detection and segmentation models are pivotal for enhancing agricultural studies and innovations through image processing \& analysis \cite{dhanya2022deep, tian2020computer, mavridou2019machine} and automation \& robotics \cite{thakur2023extensive, jha2019comprehensive}, particularly in complex environments such as apple orchards \cite{meng2025yolov10, sapkota2024integrating, sapkota2024zero}. The evolution of the YOLO (You Only Look Once) series marks significant milestones in this journey. Since the inception of the first YOLO model in 2016 by Joseph Redmon \cite{redmon2016you}, the series has evolved significantly, culminating in the release of YOLO11 in September 2024 by Ultralytics, based in California, United States. The YOLO series, spanning from YOLOv1 to YOLO11, has seen various enhancements, but only select versions such as YOLOv5, YOLOv8, and YOLO11 offer instance segmentation capabilities. These three versions were developed by Ultralytics and are documented in their respective repositories (https://github.com/ultralytics/yolov5, https://github.com/ultralytics/ultralytics, https://docs.ultralytics.com/models/yolo11/). Other variants were contributed by different authors, including YOLOv2 by Redmon et al. \cite{redmon2017yolo9000}, YOLOv3 by Farhadi et al. \cite{farhadi2018yolov3}, YOLOv4 by Bochkovskiy et al. \cite{bochkovskiy2020yolov4}, YOLOv6 by Li et al. \cite{li2022yolov6}, YOLOv7 and YOLOv9 by Wang et al. \cite{wang2023yolov7, wang2024yolov9}, and YOLOv10 by Wang et al. \cite{wang2024yolov10}. 

Each iteration of the YOLO series has progressively outperformed its predecessors in benchmark datasets, with YOLO11 representing the latest development capable of advanced instance segmentation in RGB images \cite{sapkota2024yolov10, sapkota2024comprehensive}. This capability makes it particularly suited for complex imaging tasks such as those encountered in agriculture \cite{sapkota2024comprehensive, ranjanYOLO11, sapkota2024compares}. Previously, we extensively compared YOLOv8 against models like Mask R-CNN, particularly focusing on their instance segmentation capabilities within complex orchard environments \cite{sapkota2024comparing}. The findings indicated that YOLOv8 excelled in both accuracy and processing speed, making it a preferred choice for agricultural applications involving the segmentation of green fruits. 

Despite its successes, the release of YOLO11 necessitates a thorough evaluation of its advancements and effectiveness relative to YOLOv8. The need to compare YOLO11 with YOLOv8 stems from the ongoing pursuit of more accurate, faster, and computationally efficient models that can better address the real-world challenges of agricultural imaging. 

In orchards where green fruits often blend into the foliage, the challenges of effective detection and segmentation are critical for advancing several practical applications \cite{sapkota2024immature, sun2019recognition, liu2022accurate}. These include precision agriculture, where automated monitoring of crop health and ripeness stages facilitates timely harvesting, and yield estimation, which relies on accurate fruit count and size measurements from images to aid in supply chain and market planning. Additionally, early detection of pest attacks or diseased fruits is crucial for allowing targeted treatment interventions \cite{sapkota2024immature, he2022fruit}. This capability is also instrumental in guiding autonomous robotic systems that harvest ripe fruits without causing damage \cite{wang2021channel, kang2019fruit, jia2020detection}. Furthermore, optimizing the use of resources such as water, fertilizers, and pesticides can be achieved by accurately detecting fruit loads and assessing plant health \cite{corceiro2023methods}. Additionally, high-throughput plant phenotyping, which evaluates traits related to fruit size, shape, and color distribution, also benefits significantly from advancements in fruit detection technologies in orchards \cite{jiang2020convolutional, james2022high, ninomiya2022high}.

The objective of this study is to compare the capabilities of YOLO11 and YOLOv8, along with their respective configurations, in performing instance segmentation of immature green fruits within a commercial orchard environment. Specifically, this analysis aims to discern how each model and its configurations handle the segmentation of fruits under two distinct categories: occluded and non-occluded. By examining these capabilities, the study seeks to identify which model configuration delivers the most accurate and efficient instance segmentation results in a challenging green fruit environment of commercial Scifresh apple orchard. The specific objectives of this study are :
\begin{itemize}
    \item \textbf{Data Acquisition}: To collect consumer-grade camera-based RGB images of immature fruitlets through a robotized platform for assembling a comprehensive dataset that is precisely timed before thinning, during extensive crop care requirements.
    \item \textbf{Dataset Preparation}: To manually prepare the occlusion and non-occlusion classes of datasets through the heavy effort of annotations to test the deep learning performance.
    \item \textbf{Deep Learning Based Instance Segmentation}: Implement YOLO11 and YOLOv8, along with their all configurations under identical hyperparameter settings for instance segmentation to identify distinct characteristics and stages of crop growth.
    \item \textbf{Performance Evaluation}: Assessing the models in terms of precision, recall, F1-scores, mAP@50, GFLOPS, number of parameters, training epochs, convolution layers and inference speed to determine their efficacy in real-world agricultural applications.
\end{itemize}

\section{YOLOv8 and YOLO11 Overview}
\begin{figure}[ht]
\centering
\includegraphics[width=0.84\linewidth]{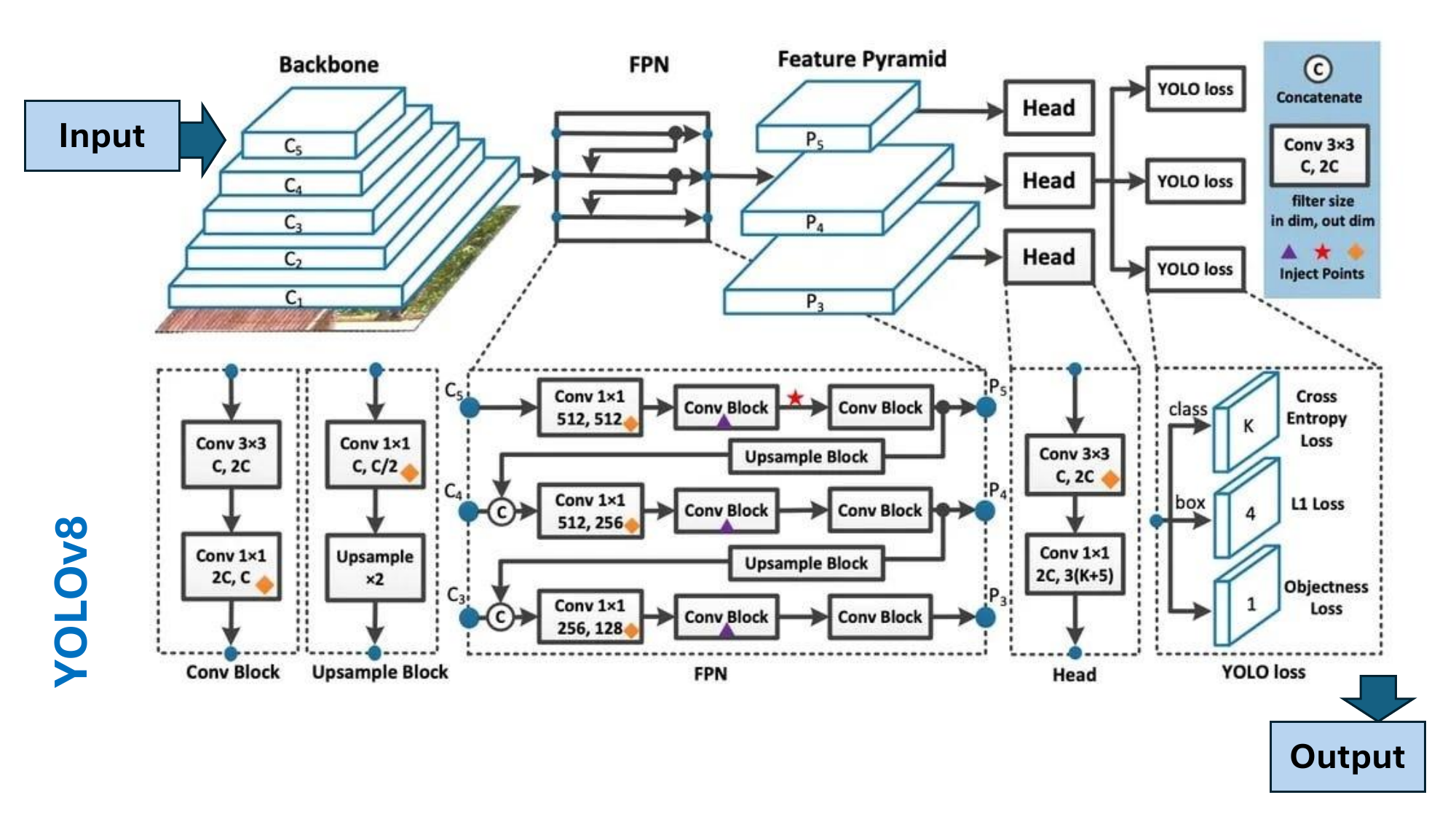}
\caption{Architecture diagram of YOLOv8 algorithm: YOLOv8 advances real-time object detection with its innovative backbone and anchor-free Ultralytics head, optimizing detection accuracy and speed across various tasks (primary image source: https://yolov8.org/what-is-yolov8/).}
\label{fig:Figure1}
\end{figure}

\begin{figure}[ht]
\centering
\includegraphics[width=0.9\linewidth]{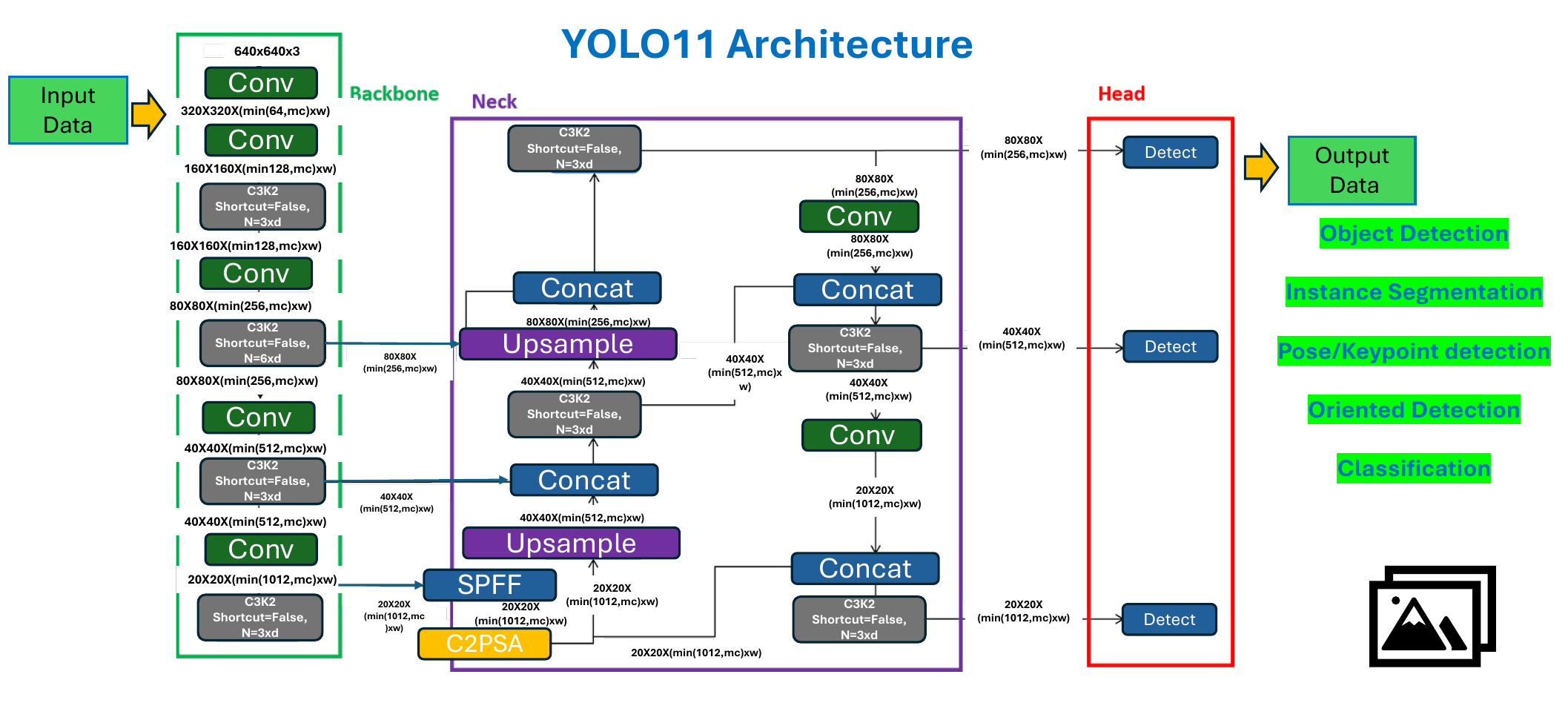}
\caption{Architecture diagram of YOLO11 algorithm: YOLO11 builds on YOLOv8's architecture  with refined detection and segmentation efficiency and accuracy over benchmark dataset \cite{sapkota2024yolov10}}
\label{fig:Figure2}
\end{figure}

YOLOv8 and YOLO11, subsequent developments by Ultralytics following their initial foray into the YOLO series with YOLOv5, represent the forefront of real-time object detection technology. These models extend capabilities across detection, segmentation, and pose estimation, unalike other YOLO variants. YOLOv8 introduced an advanced backbone and neck architecture as shown in Figure \ref{fig:Figure1}, which enhances feature extraction and improves object detection accuracy significantly. YOLOv8 also employed an anchor-free split Ultralytics head, optimizing the accuracy-speed tradeoff crucial for real-time applications and making it ideal for a range of tasks including traditional object detection and instance segmentation. On the other hand, YOLO11(Figure \ref{fig:Figure2}) is built upon the foundation laid by YOLOv8, incorporating further optimizations that boost processing speed and model efficiency \cite{sapkota2024comprehensive, sapkota4941582synthetic}. It introduces refined architectural designs and training methodologies, achieving greater accuracy with fewer parameters compared to its predecessors. Both YOLO11 and YOLOv8 models support a variety of operational modes like inference, validation, training, and export, making them adaptable across different environments from edge devices to cloud systems \cite{sapkota2024comprehensive, sapkota4941582synthetic, sapkota2024yolov10}. 


\subsection{YOLOv8 Overview}
YOLOv8 represents a significant leap in the YOLO series of real-time object detectors, pushing the boundaries of accuracy and speed for various object detection tasks. Building upon the advancements from previous versions, YOLOv8 integrates new features and optimizations, enhancing its application across a broad spectrum of tasks.

\textbf{Key Features of YOLOv8 include:}
\begin{itemize}
    \item Advanced Backbone and Neck Architectures for superior feature extraction.
    \item Anchor-free Split Ultralytics Head improving accuracy and detection efficiency.
    \item Optimized Accuracy-Speed Tradeoff, making it suitable for real-time applications.
    \item A diverse range of Pre-trained Models to cater to specific performance requirements.
\end{itemize}

YOLOv8 supports a wide variety of tasks, including detection, instance segmentation, pose estimation, and more, each optimized for high performance and accuracy.

\textbf{Performance Metrics:}
YOLOv8 models are designed to deliver cutting-edge performance, with impressive metrics on benchmark datasets like COCO and ImageNet, YOLOv7, YOLOv6, YOLOv6 and so on as dipicted in Figure \ref{fig:performancemetricsYOLO11}a.

\subsection{YOLO11 Overview}
YOLO11 is the latest iteration in the Ultralytics YOLO series, further enhancing the capabilities introduced by YOLOv8. YOLO11 is engineered to deliver unmatched accuracy, efficiency, and speed, making it a formidable tool for a range of computer vision tasks. The YOLO11 model utilizes enhanced training techniques that have led to improved results on benchmark datasets. Notably, YOLO11m achieved a mean Average Precision (mAP) score of 95.0\% on the COCO dataset while utilizing 22\% fewer parameters compared to YOLOv8m, demonstrating greater efficiency without compromising accuracy. With an average inference speed that is 2\% faster than YOLOv10, YOLO11 is optimized for real-time applications, ensuring quick processing even in demanding environments.

\begin{figure}[ht]
\centering
\includegraphics[width=0.8\linewidth]{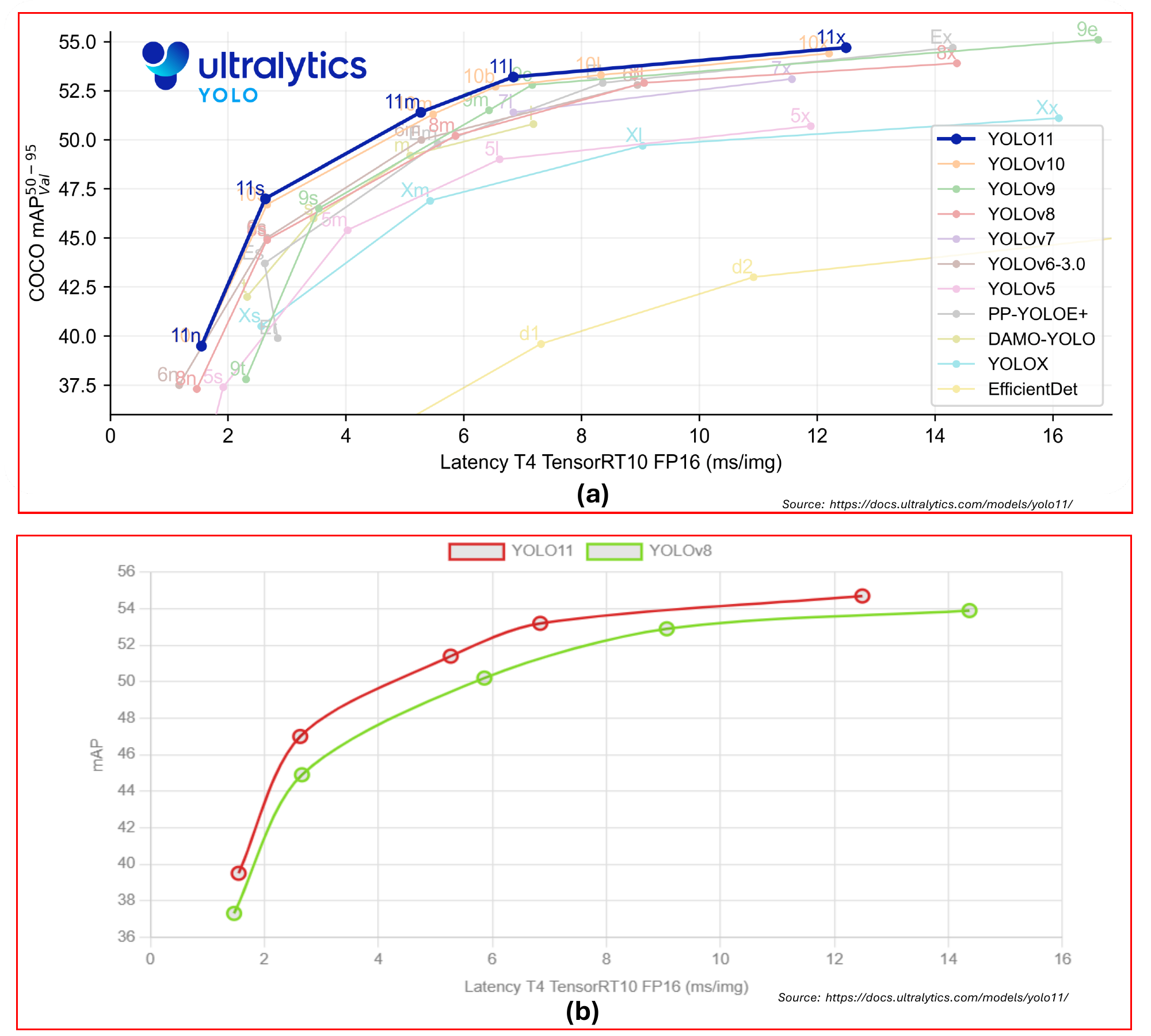}
\caption{\textbf{YOLO11 and YOLOv8 performance results on benchmark datasets, showcasing YOLO11's superior accuracy and latency against predecessors and competitors. YOLO11 achieves the highest mAP scores at varied latency levels, highlighting its efficiency in real-time applications:}(a) Performance comparison of YOLO models on the COCO dataset, ; (b) Performance comparison between YOLO11 and YOLOv8 on the COCO dataset}
\label{fig:performancemetricsYOLO11}
\end{figure}

\textbf{Advancements in YOLO11:}
\begin{itemize}
    \item Enhanced Feature Extraction: Utilizes an advanced backbone and neck architecture, enabling more precise object detection.
    \item Optimized for Efficiency and Speed: Features refined architectural designs and training methodologies for faster processing.
    \item Greater Accuracy with Fewer Parameters: Achieves higher mAP scores on the COCO dataset with fewer parameters compared to YOLOv8m.
    \item Broad Adaptability: Can be deployed across various platforms, including edge devices and cloud systems.
\end{itemize}

YOLO11 supports an extensive range of tasks from simple object detection to complex oriented object detection and segmentation, providing flexible deployment options from inference to export.

\textbf{Performance Metrics:}
The model exhibits superior performance metrics on benchmark datasets compared to YOLOv10, YOLOv9, YOLOv8, YOLOv7, YOLOv6, PP-YOLOE+, DAMO-YOLO, YOLOX and EfficientDet as depicted in Figure \ref{fig:performancemetricsYOLO11}a and specific comparison between YOLO11 and YOLOv8 is shown in \ref{fig:performancemetricsYOLO11}b.
\section{Methodology}
\subsection{Data Acquisition}
\begin{figure}[ht]
\centering
\includegraphics[width=0.87\linewidth]{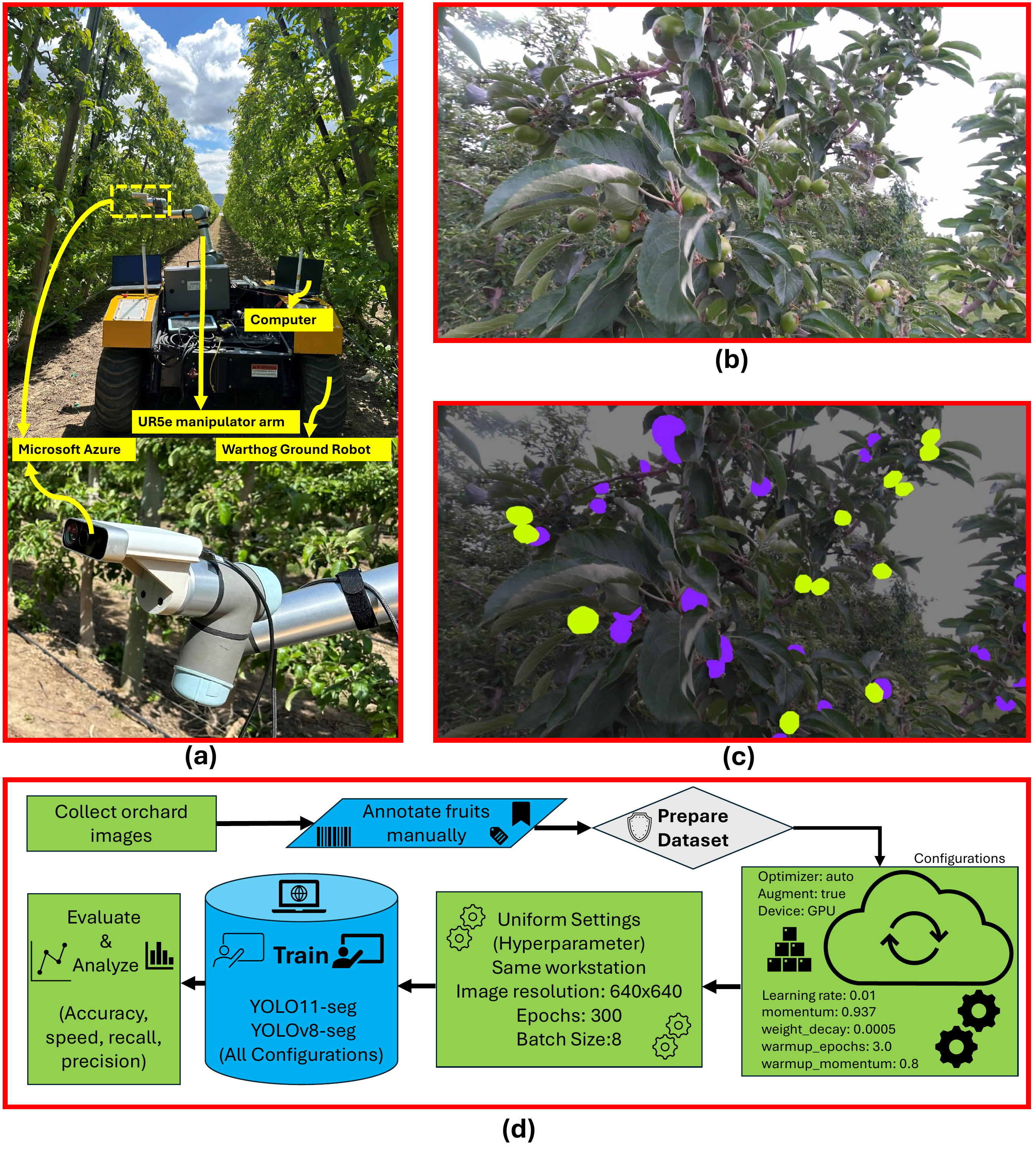}
\caption{Data acquisition, annotation and training deep learning process diagram: (a) Robotic imaging platform equipped with a Microsoft Azure Kinect DK sensor on a UR5e manipulator arm, navigating through an orchard; (b) Example of RGB images captured showing complex backgrounds; (c) Manual annotation of occluded and non-occluded immature green fruits; (d) Process diagram: prepared dataset split into training, testing, and validation sets.}
\label{fig:Figure3}
\end{figure}

The study was conducted in a commercial orchard located in Prosser, Washington State, USA, planted with 'Scifresh' apple variety. The orchard was structured with tree rows planted 10 feet apart and a spacing of 3 feet between each tree, with the tree height maintained at approximately 10 feet. A robotic imaging platform equipped with a Microsoft Azure Kinect DK sensor (Microsoft Azure, Washington State, USA), mounted on a UR5e industrial manipulator arm (Universal Robotics, Odense S, Denmark), navigated through the orchard on a Warthog ground robot (Clearpath Robotics, Ontario, Canada). This setup captured high-resolution RGB images during the immature fruitlet stage, as depicted in Figure \ref{fig:Figure3}a and Figure \ref{fig:Figure3}b.
\subsection{Data Preprocessing and Preparation}
A total of 991 images were collected and manually annotated to classify immature fruitlets as either occluded or non-occluded based on their visibility. This process, shown in Figure \ref{fig:Figure3}c, involved labeling fully visible fruits as 'non-occluded apples' and those obscured by foliage or other orchard elements as 'occluded apples'. The annotated dataset was then formatted for compatibility with YOLO11 and YOLOv8 architectures and split into training, testing, and validation sets in an 8:1:1 ratio, with detailed preparations illustrated in Figure \ref{fig:Figure3}d.  All the data annotations and preparation for this study was conducted on Roboflow(roboflow.com). 

\subsection{YOLO11 and YOLOv8 Training for Instance Segmentation}
Both the YOLO11 and YOLOv8 models were trained on a workstation with an Intel Xeon® W-2155 CPU @ 3.30 GHz x20 processor, NVIDIA TITAN Xp Collector's Edition/PCIe/SSE2 graphics card, 31.1 GiB memory, and Ubuntu 16.04 LTS 64-bit operating system.  Likewise, both models underwent identical training protocols to ensure a balanced evaluation. The dataset comprised manually annotated images of immature green fruits in varying states of occlusions processed under a uniform configuration to facilitate consistent learning across both model architectures. Each model utilized pre-trained weights to accelerate the convergence on the specialized task of the immature green fruit segmentation, harnessing previously learned features relevant to the dataset.

For both YOLO11-seg and YOLOv8-seg models, the training was conducted over 300 epochs to thoroughly adapt the models to the complexities of the orchard environment. The deterministic nature of the training (deterministic: true), combined with a fixed random seed (seed: 0) was ensured for reproducibility by eliminating stochastic elements in the training process. This approach was crucial for maintaining uniformity in model evaluation under a similar hyper-parameter environment.

Additionally, RGB image inputs were standardized to a resolution of 640x640 pixels (imgsz: 640), to balance detail with processing efficiency, suitable for the intricate details required in distinguishing occluded immature green fruits. The batch size was set at eight (batch: 8), optimizing the balance between memory constraints and batch performance. The models applied both horizontal flipping (fliplr: 0.5) and minimal rotation (degrees: 0.0), mirroring real-world variations in fruit orientation and positioning, which were critical for the robust segmentation of objects in practical applications.

The Intersection Over Union (IOU) threshold was established at 0.7 (iou: 0.7), to ensure high accuracy in the segmentation masks by requiring a substantial overlap between the predicted and actual labels. This high threshold was pivotal for precise segmentation of occluded fruits, where accurate delineation is essential for subsequent agricultural processes. The training setup also included a stringent control over model complexity and overfitting through a modest learning rate (lr0: 0.01) and weight decay (weight decay: 0.0005).

\subsection{Performance Evaluation of YOLO11 and YOLOv8 for Instance Segmentation of Immature Green Fruits}
The effectiveness of both YOLO11 and YOLOv8 models in segmenting immature green fruits, categorized into occluded and non-occluded apples, was evaluated using comprehensive metrics such as Mean Intersection over Union (MIoU), Average Precision (AP), Mean Average Precision (mAP), Mean Average Recall (mAR), and F1-score. The MIoU, also recognized as the Jaccard index, gauges the precision of segmentation by comparing the overlap and union areas of the predicted and actual targets:

\begin{equation}
MIoU = \frac{Area_{Overlap}}{Area_{Union}} = \frac{TP}{FP + TP + FN}
\end{equation}

where:
\begin{itemize}
    \item $TP$ (True Positives) counts the immature green apples accurately identified by the model.
    \item $FP$ (False Positives) indicates the immature green non-apples erroneously marked as apples.
    \item $FN$ (False Negatives) denotes the immature green apples that were overlooked by the model.
\end{itemize}

Precision is defined as the accuracy of positive predictions:

\begin{equation}
Precision = \frac{TP}{TP + FP}
\end{equation}

Recall measures the model's ability to detect all actual positives:

\begin{equation}
Recall = \frac{TP}{TP + FN}
\end{equation}

The F1-score, which harmonizes precision and recall, is crucial for models where an equilibrium between these metrics is sought:

\begin{equation}
F1\text{-}Score = 2 \times \frac{(Precision \times Recall)}{Precision + Recall}
\end{equation}

AP and mAP serve to quantify the model’s effectiveness across different threshold levels, with mAP averaging the AP for all classes to provide a global perspective on the model’s overall segmentation prowess. This rigorous evaluation framework facilitates a detailed comparison of YOLO11 and YOLOv8's capabilities in accurately detecting and segmenting both occluded and non-occluded immature green apples in challenging orchard environments.

\subsection{Evaluation of Parameters, GFLOPs, and Layers in YOLO11 and YOLOv8 Configurations} The architectural complexity and computational efficiency of the YOLO11 and YOLOv8 object detection and segmentation algorithms were systematically assessed by examining three critical aspects: the number of parameters, GFLOPs (Giga Floating Point Operations per Second), and the count of convolutional layers utilized in each configuration. These factors are essential indicators of a model's potential performance and operational demands.

Parameters, representing the total count of trainable elements within the model, were evaluated to understand the models' complexity and memory requirements:

\begin{equation} Parameters_{Model} = \text{Total trainable weights and biases} \end{equation}

GFLOPs were calculated to estimate the computational load during the inference phase, providing insights into the models’ efficiency and speed:

\begin{equation} GFLOPs = \frac{\text{Total floating-point operations}}{10^9} \text{ per image} \end{equation}

The number of convolutional layers, which directly influences feature extraction capabilities and depth of the network, was recorded to measure the architectural depth:

\begin{equation} Layers_{Convolutional} = \text{Total number of convolutional layers in the model} \end{equation}

\section{Results and Discussion}
\begin{figure}[ht]
\centering
\includegraphics[width=0.9\linewidth]{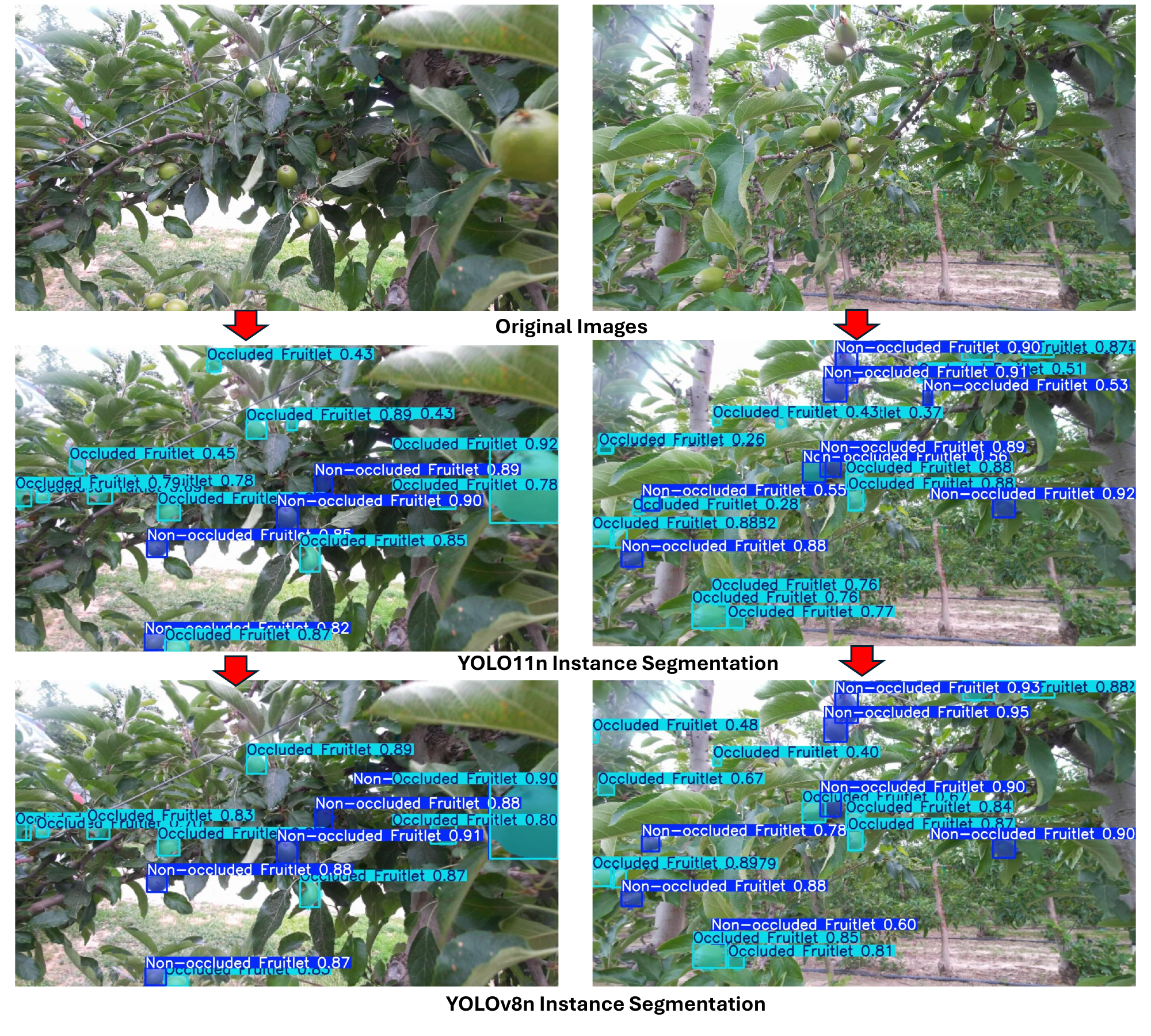}
\caption{Examples showing YOLO11n and YOLOv8n's results in segmenting immature green fruits in a commercial Scilate orchard. The top row displays original RGB images, the middle row illustrates YOLO11n's segmentation results, and the bottom row showcases YOLOv8n's performance on the same images.}
\label{fig:Figure7}
\end{figure}

The examples of instance segmentation for occluded and non-occluded immature green fruits are illustrated in Figure \ref{fig:Figure7}, which presents a visual comparison of instance segmentation results for immature green fruits using YOLO11-seg and YOLOv8-seg models. Both models demonstrated impressive segmentation capabilities, with YOLO11n-seg achieving the highest precision overall. In terms of mAP@50, YOLO11l-seg recorded the highest value for occluded immature green apple segmentation, reaching a score of 0.81. Among the YOLOv8 variants, YOLOv8m-seg exhibited the highest precision at 0.824, and YOLOv8l-seg achieved a notable mAP@50 of 0.807, underscoring their effectiveness in distinguishing between occluded and non-occluded fruits in complex orchard environments.

The rest of this section is organized into five subsections as \textit{Evaluation Metrics for YOLO11 and YOLOv8: Precision, Recall, and mAP@50}, \textit{Evaluation of Parameters, GFLOPs, and Layers used in training YOLO11 and YOLOv8}, \textit{Evaluation of Training Time}, \textit{Evaluation of Image Processing Speeds}, and \textit{Discussion of Findings}.

\subsection{Evaulation Metrics for YOLO11 and YOLOv8 : Precision, Recall and mAP@50}
In the assessment of instance segmentation capabilities across YOLO11 and YOLOv8 models, performance metrics delineate a clear differentiation in effectiveness by category: All, non-occluded, and occluded fruitlets. YOLO11n-seg led the segmentation precision for all categories with a significant score of 0.831, underscoring its robustness in comprehensive fruit detection. In the analysis of non-occluded and occluded fruitlets, YOLO11m-seg and YOLO11l-seg models achieved highes scores as 0.851 and 0.829 respectively. This precision is critical in agricultural applications where accurate instance segmentation influences operational efficiency, such as in real-time size estimation of immature green fruits as experimented in our previous study \cite{sapkota2024immature}. Furthermore, YOLO11x-seg emerged as a leader in mask recall across all categories with an impressive score of 0.815, while YOLO11m-seg demonstrated superior performance in identifying non-occluded fruitlets with a recall of 0.858. YOLOv8x-seg marked its excellence in the occluded category, achieving the highest mask recall of 0.800. 

However, in terms of box detection metrics, the YOLOv8n models surpassed the YOLO11 models in instance segmentation for the non-occluded immature green fruits class with a score of 0.851. Meanwhile, the YOLO11l configuration outperformed all other model configurations in instance segmentation for the "All" and "Occluded Fruitlet" classes, achieving scores of 0.842 and 0.852, respectively. The detailed values of metrics for box detection and mask instance segmentation are presented in Table \ref{tab:metrics} as precision and recall values of YOLO11 and YOLOv8 and their configurations.

\begin{table}[ht]
\centering
\caption{\textbf{Comparative performance metrics of YOLO11 and YOLOv8 models for instance segmentation of immature green apples in orchards. Metrics are presented for Box and Mask accuracies, divided into overall, non-occluded, and occluded fruitlet categories, showcasing precision and recall values across ten model configurations. The outperforming model in each category for each column is highlighted in bold}}
\label{tab:metrics}
\begin{tabular}{@{}c|ccc|ccc@{}}
\toprule
\textbf{Model} & \multicolumn{3}{c|}{\textbf{Box}} & \multicolumn{3}{c}{\textbf{Mask}} \\
\cmidrule(lr){2-4} \cmidrule(lr){5-7}
 & \textbf{All} & \textbf{Non-occluded} & \textbf{Occluded} & \textbf{All} & \textbf{Non-occluded} & \textbf{Occluded} \\
\midrule
\textbf{Precision} \\
YOLO11n         &0.809 & 0.809& 0.81 & 0.795&0.801 &0.789 \\
YOLO11s         &0.832 &0.845 &0.819 &0.817 &0.836 &0.798 \\
YOLO11m         & 0.832&0.843 &0.816 &0.828 & \textbf{0.851} &0.804 \\
YOLO11l         & \textbf{0.842}& 0.831& \textbf{0.852} & \textbf{0.831} &0.832 & \textbf{0.829} \\
YOLO11x         & 0.802& 0.829&0.776 &0.786 &0.825 &0.747 \\
YOLOv8n         & 0.83& \textbf{0.851} &0.81 &0.819 &0.847 &0.791 \\
YOLOv8s         & 0.815&0.816 &0.814 &0.8 &0.812 &0.789 \\
YOLOv8m         &0.824 &0.832 &0.816 &0.805 &0.825 &0.784 \\
YOLOv8l         &0.819 &0.831 &0.807 & 0.806&0.829 &0.782 \\
YOLOv8x         &0.787 &0.819 &0.755 &0.777 &0.815 &0.738 \\
\midrule
\textbf{Recall} \\
YOLO11n         &0.685 &0.745 &0.624 &0.668 &0.734 &0.602 \\
YOLO11s         &0.815 &0.854 &0.776 &0.793 &0.839 &0.747 \\
YOLO11m         &0.829 & \textbf{0.865} &0.793 &0.812 & \textbf{0.858} &0.765 \\
YOLO11l         &0.8 &0.846 &0.755 &0.783 &0.839 &0.726 \\
YOLO11x         &0.832 &0.853 &0.811 & \textbf{0.815} &0.849 &0.781 \\
YOLOv8n         &0.785 &0.835 &0.735 &0.773 &0.83 &0.716 \\
YOLOv8s         &0.801 &0.843 &0.759 &0.787 &0.839 &0.735 \\
YOLOv8m         &0.821 &0.858 &0.783 &0.799 &0.85 &0.749 \\
YOLOv8l         &0.824 &0.842 &0.806 &0.798 &0.83 &0.766 \\
YOLOv8x         & \textbf{0.845} &0.85 & \textbf{0.841} &0.811 &0.822 & \textbf{0.8} \\
\bottomrule
\end{tabular}
\end{table}

\begin{table}[ht]
\centering
\caption{\textbf{mAP@50 for Box and Mask categories across YOLO11 and YOLOv8 configurations}}
\label{tab:map50}
\begin{tabular}{@{}llcccccc@{}}

Model & Category & \multicolumn{3}{c}{YOLO11} & \multicolumn{3}{c}{YOLOv8} \\ 
      &          & All   & Non-occluded & Occluded & All   & Non-occluded & Occluded \\ \midrule
{n-seg} & Box & 0.752 & 0.801 & 0.702 & 0.843 & 0.894 & 0.792 \\
                       & Mask & 0.736 & 0.789 & 0.682 & 0.829 & \textbf{0.889} & 0.769 \\
{s-seg} & Box & 0.864 & 0.893 & 0.835 & 0.855 & 0.877 & 0.823 \\
                       & Mask & 0.837 & 0.878 & 0.795 & 0.840 & 0.885 & 0.794 \\
{m-seg} & Box & \textbf{0.876} & \textbf{0.908} & 0.844 & 0.852 & 0.866 & 0.839 \\
                       & Mask & \textbf{0.860} & \textbf{0.909} & \textbf{0.810} & 0.829 & 0.862 & 0.797 \\
{l-seg} & Box & 0.873 & 0.898 & \textbf{0.847} & \textbf{0.873} &  \textbf{0.898} & \textbf{0.847} \\
                       & Mask & 0.851 & 0.893 & 0.809 &  \textbf{0.848} & 0.889 & \textbf{0.807} \\
{x-seg} & Box & 0.875 & 0.905 & 0.845 & 0.865 & 0.891 & 0.838 \\
                       & Mask & 0.852 & 0.904 & 0.800 & 0.841 & 0.882 & 0.800 \\

\end{tabular}
\end{table}

In the evaluation of mean average precision at a 50\% intersection over union threshold (mAP@50), focusing on instance segmentation performance, YOLO11m-seg consistently delivered superior outcomes across all categories. For the "All" class, this model not only achieved the highest mAP@50 in box detection at 0.876 but also excelled in mask segmentation with an impressive score of 0.860. In the non-occluded fruitlet category, YOLO11m-seg maintained its lead, registering the highest scores for both box and mask mAP@50, marked at 0.908 and 0.909, respectively. This underscores the model's efficacy in discerning clearly visible fruits without occlusion. For the occluded fruitlet class, which presents more complex segmentation challenges due to partial visibility among foliage, there was a tie in box mAP@50 between YOLO11l-seg and YOLOv8l-seg, both scoring 0.847. However, YOLO11m-seg outstripped its counterparts in mask mAP@50 with a score of 0.810, highlighting its robust capability in accurately segmenting fruits obscured by surrounding elements. Detailed values of mAP@50 for this study are presented in Table \ref{tab:map50}.

\begin{figure}[ht]
\centering
\includegraphics[width=0.95\linewidth]{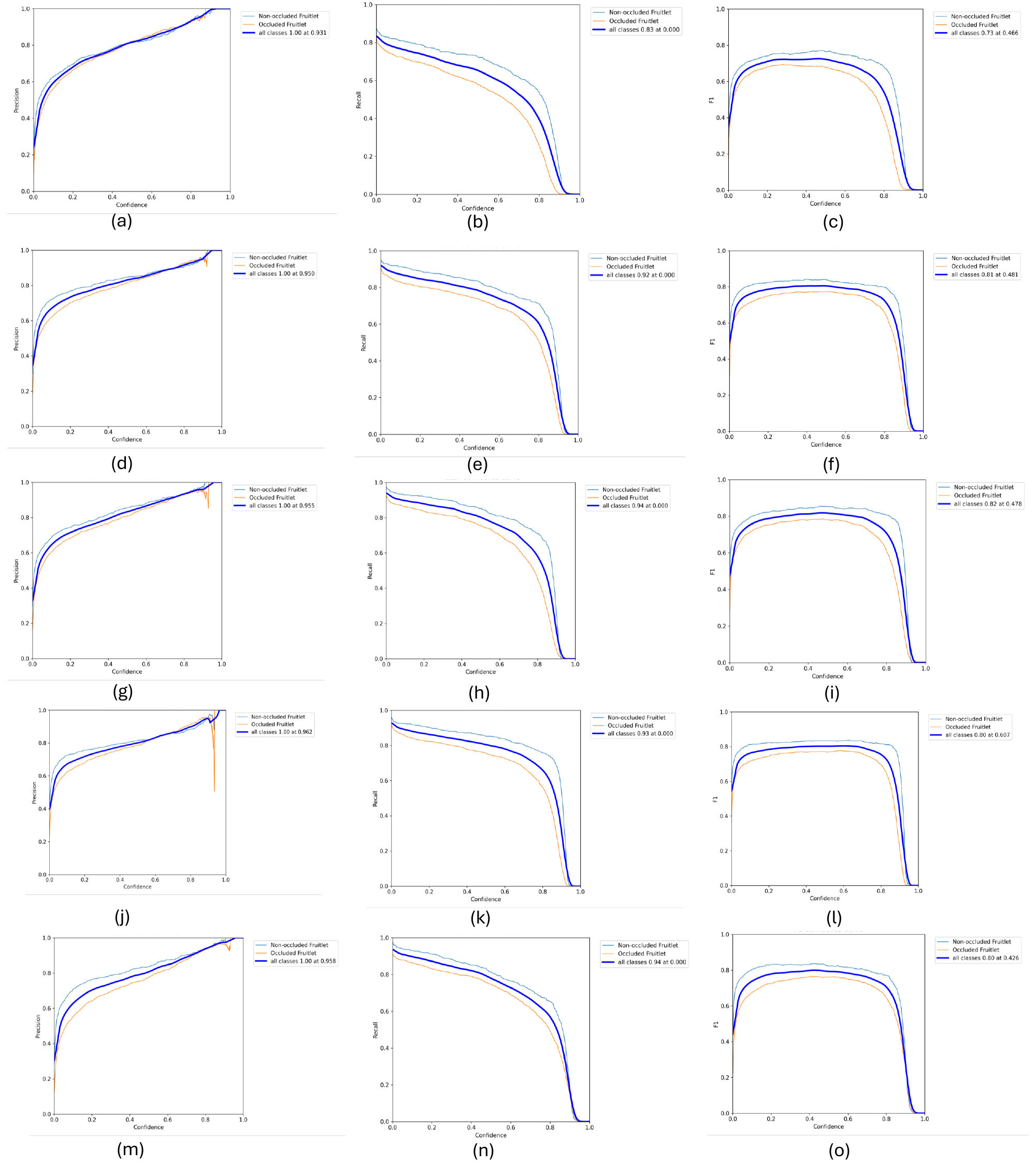}
\caption{Precision, Recall and F1 score curves for Mask Results of YOLO11: (a-c) Precision, recall, and F1-score for YOLO11n; (d-f) same metrics for YOLO11s; (g-i) for YOLO11m; (j-l) for YOLO11l; and (m-o) for YOLO11x, in instance segmentation immature green fruits for occluded and non-occluded class.}
\label{fig:Figure9}
\end{figure}

The precision, recall, and F1-score metrics for all configurations of YOLO11 are depicted in Figure \ref{fig:Figure9}. Figures \ref{fig:Figure9}a, \ref{fig:Figure9}b, and \ref{fig:Figure9}c illustrate the precision, recall, and F1-score for YOLO11n, respectively. Figures \ref{fig:Figure9}d, \ref{fig:Figure9}e, and \ref{fig:Figure9}f present these metrics for YOLO11s, while figures \ref{fig:Figure9}g, \ref{fig:Figure9}h, and \ref{fig:Figure9}i do the same for YOLO11m. For YOLO11l, these metrics are shown in figures \ref{fig:Figure9}j, \ref{fig:Figure9}k, and \ref{fig:Figure9}l, and for YOLO11x in figures \ref{fig:Figure9}m, \ref{fig:Figure9}n, and \ref{fig:Figure9}o.

\begin{figure}[ht]
\centering
\includegraphics[width=0.8\linewidth]{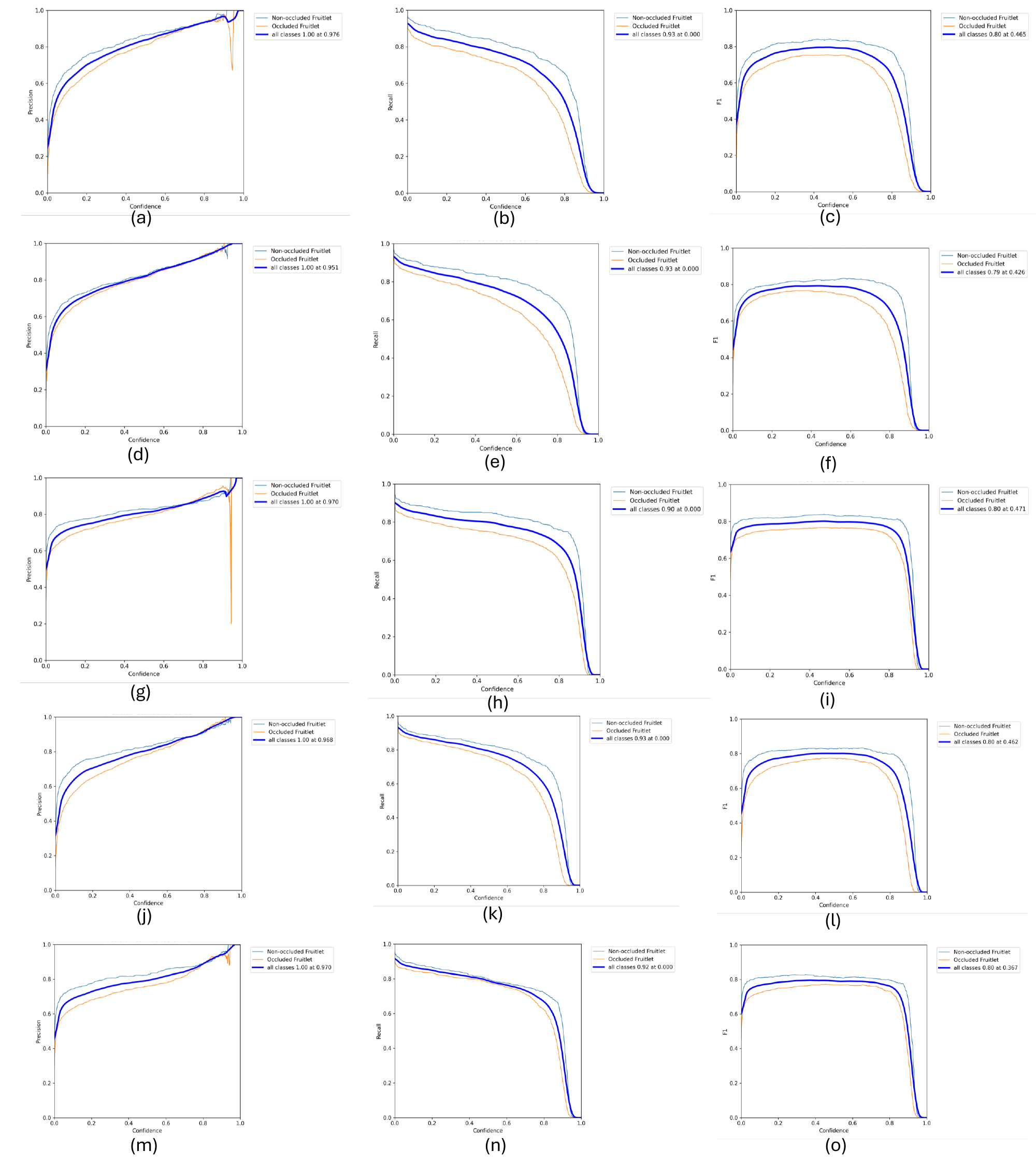}
\caption{Precision, Recall and F1 score curves for Mask Results of YOLOv8 : (a-c) Precision, recall, and F1-score for YOLOv8n; (d-f) the same metrics for YOLOv8s; (g-i) for YOLOv8m; (j-l) for YOLOv8l; and (m-o) for YOLOv8x, demonstrating each variant's effectiveness in segmenting occluded and non-occluded immature green fruits.}
\label{fig:Figure10}
\end{figure}

Likewise,the precision, recall, and F1-score metrics for all configurations of YOLOv8 are detailed in Figure \ref{fig:Figure10}. Figures \ref{fig:Figure10}a, \ref{fig:Figure10}b, and \ref{fig:Figure10}c display the precision, recall, and F1-score for YOLOv8n, respectively. Figures \ref{fig:Figure10}d, \ref{fig:Figure10}e, and \ref{fig:Figure10}f showcase these metrics for YOLOv8s, while figures \ref{fig:Figure10}g, \ref{fig:Figure10}h, and \ref{fig:Figure10}i present them for YOLOv8m. For YOLOv8l, these metrics are illustrated in figures \ref{fig:Figure10}j, \ref{fig:Figure10}k, and \ref{fig:Figure10}l, and for YOLOv8x in figures \ref{fig:Figure10}m, \ref{fig:Figure10}n, and \ref{fig:Figure10}o

\subsection{Evaluation of Parameters, GFLOPs and Layers used in training YOLO11 and YOLOv8}
\begin{figure}[ht]
\centering
\includegraphics[width=0.95\linewidth]{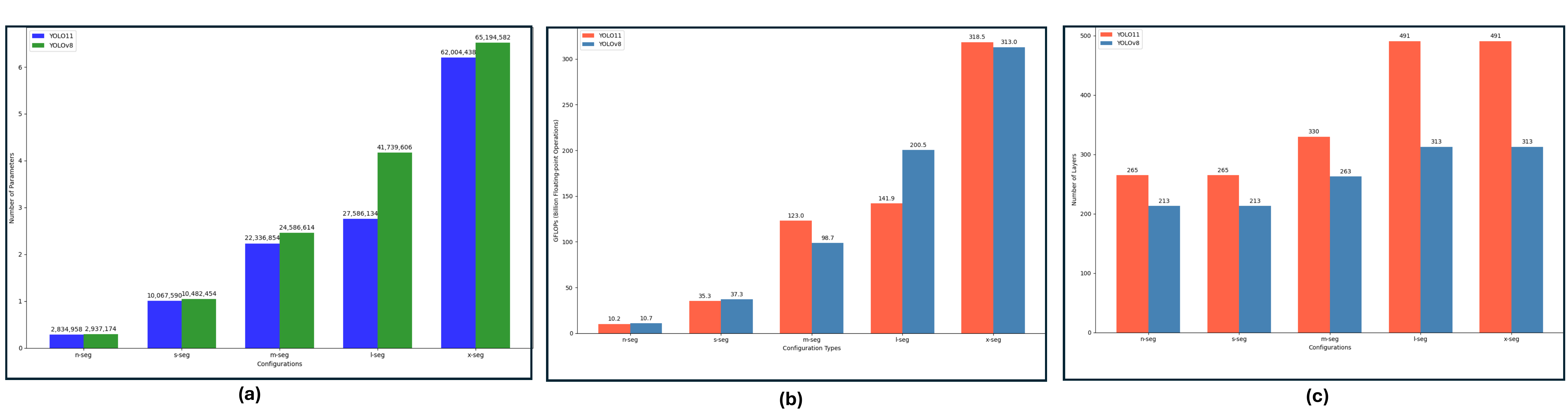}
\caption{Showing a comparative analysis of YOLO11 and YOLOv8 configurations for detecting immature green fruitlets in two classes. Subfigure (a) illustrates the number of parameters, (b) shows the GFLOPs, and (c) details the layers used in each model configuration. }
\label{fig:Figure5}
\end{figure}

In the detailed evaluation of parameter utilization across the YOLO11 and YOLOv8 segmentation models, distinct patterns emerged, highlighting the computational intricacies inherent in each model configuration. The YOLO11 models demonstrated a parameter range from 2.83 million in the 'YOLO11n-seg' configuration, which is the least, to a substantial 62.00 million in the 'YOLO11x-seg' configuration, indicating the highest parameter usage within the YOLO11 series. Conversely, the YOLOv8 series started slightly higher with 2.94 million parameters for 'YOLOv8n-seg' and peaked at 65.19 million parameters for 'YOLOv8x-seg'. These figures underscore the varying complexity and potential computational load, with further details visualized in Figure \ref{fig:Figure5}a showing a bar diagram comparing parameter counts across all configurations of YOLO11 and YOLOv8 object detection and instance segmentation algorithm. 

In terms of computational complexity measured by GFLOPs, the YOLO11 and YOLOv8 segmentation configurations demonstrate a significant range, adapting to diverse operational requirements. For YOLO11, the YOLO11x-seg configuration demands the highest computational resources, clocking in at 318.5 GFLOPs, which supports its capacity for handling more intricate segmentations. The YOLO11n-seg model is the most computationally efficient within this series, utilizing only 10.2 GFLOPs, designed for environments where lower computational load is crucial. In contrast, the YOLOv8 series shows a similar trend where YOLOv8x-seg reaches up to 313 GFLOPs, providing extensive computational power for complex tasks, while YOLOv8n-seg remains the least demanding with 10.7 GFLOPs. This spread in computational needs across models is further elaborated in Figure \ref{fig:Figure5}b, which illustrates the GFLOPs distribution, reflecting the trade-offs between performance and computational efficiency in different segmentation scenarios.

In the segmentation configurations of YOLO11 and YOLOv8, the architectural depth is adapted to the complexity of the detection and segmentation tasks. Within the YOLO11 series, YOLO11l-seg and YOLO11x-seg models incorporate the highest number of convolutional layers, totaling 491 layers each, which provides substantial discriminative capabilities for complex segmentation challenges. Conversely, YOLO11n-seg and YOLO11s-seg utilize the fewest layers, with 265 layers each, indicating a design optimized for faster inference with less computational overhead. Similarly, for YOLOv8, the YOLOv8l-seg and YOLOv8x-seg configurations utilize 313 layers, while YOLOv8n-seg and YOLOv8s-seg use 213 layers. Detailed comparisons of layer utilization across all configurations are visually depicted in Figure \ref{fig:Figure5}c, highlighting the strategic layer distribution in response to varying segmentation demands.

\begin{figure}[ht]
\centering
\includegraphics[width=0.98\linewidth]{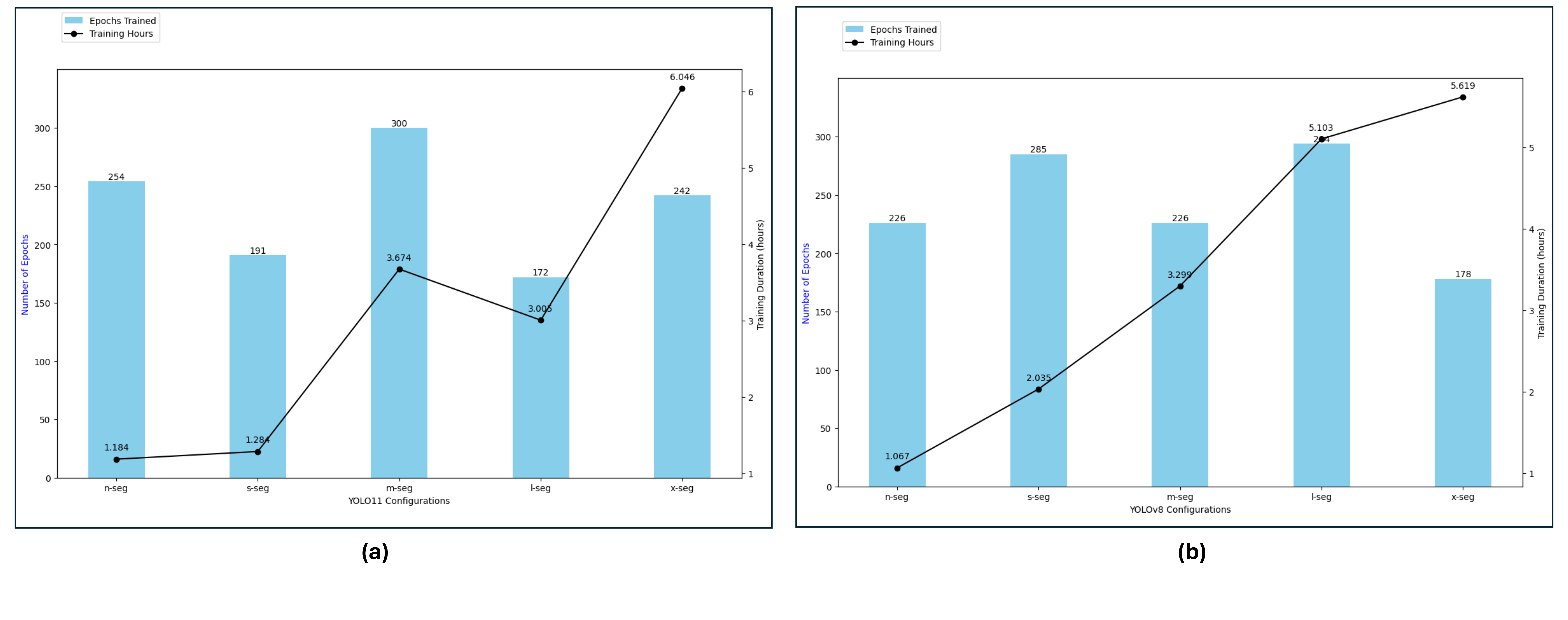}
\caption{Figure illustrates the training duration and number of epochs completed for YOLO11 (a) and YOLOv8 (b) configurations in detecting immature green fruitlets, highlighting efficiency and early stopping instances. }
\label{fig:Figure6}
\end{figure}
\subsection{Evaluation of Training Time }
The training durations and completion epochs for both YOLO11 and YOLOv8 models varied significantly, indicating differing levels of efficiency and early stopping due to lack of improvements in model performance. For YOLO11, the YOLO11x-seg configuration completed the training regimen in the shortest span of 242 epochs over 6.046 hours, indicating the most intensive learning process among its peers. In contrast, YOLO11l-seg stopped training the earliest after only 172 epochs, taking 3.005 hours, demonstrating early convergence. For YOLOv8, YOLOv8l-seg pushed close to the limit with 294 epochs over 5.103 hours, while YOLOv8x-seg ceased training after 178 epochs, consuming 5.619 hours, marking the earliest halt in this group. These dynamics illustrate the models' adaptability and efficiency in learning, with some configurations halting significantly before reaching the 300-epoch maximum, possibly due to achieving sufficient learning before the set threshold. Detailed visualizations of the training epochs and hours for each configuration are provided in Figures \ref{fig:Figure6}a and \ref{fig:Figure6}b for YOLO11 and YOLOv8, respectively, showcasing the specific training timelines and stopping points across different configurations.

\subsection{Evaluation of Image Processing Speeds}
The performance analysis of YOLO11 and YOLOv8 configurations reveals critical insights into their image processing speeds, particularly in complex environments such as the segmentation of occluded and non-occluded immature green fruitlets. A detailed examination of inference speeds indicates that Although the YOLO11 model is the latest and incorporates state-of-the-art features, YOLOv8n significantly outperforms its counterparts in the YOLO11 series in terms of image processing speed. Specifically, YOLOv8n achieves an inference speed of just 3.3 milliseconds, which is markedly faster than the fastest model in the YOLO11 series, YOLO11n, with an inference speed of 4.8 milliseconds. This superior performance of YOLOv8n underscores its efficiency in processing high-resolution images rapidly, making it exceptionally suited for real-time applications in agricultural settings where timely and accurate detection and segmentation of immature green fruits is crucial. For a comprehensive overview of the preprocessing, inference, and postprocessing speeds of all configurations within the YOLO11 and YOLOv8 series, refer to Table \ref{tab:image_processing_speeds}.

\begin{table}[ht]
\centering
\caption{\textbf{Comparison of Image Processing Speeds (in milliseconds) for YOLO11 and YOLOv8 Configurations}}
\label{tab:image_processing_speeds}
\begin{tabular}{@{}lcccccc@{}}
\toprule
\textbf{Configuration} & \multicolumn{3}{c}{\textbf{YOLO11 (ms)}} & \multicolumn{3}{c}{\textbf{YOLOv8 (ms)}} \\
\cmidrule(lr){2-4} \cmidrule(lr){5-7}
              & Preprocess & Inference & Postprocess & Preprocess & Inference & Postprocess \\ \midrule
n-seg         & 0.3        & 4.8       & 56.7        & 0.3        & 3.3       & 2.5         \\
s-seg         & 0.3        & 6.0       & 2.6         & 0.3        & 5.1       & 3.6         \\
m-seg         & 0.2        & 11.7      & 2.0         & 0.3        & 10.1      & 2.4         \\
l-seg         & 0.3        & 14.5      & 1.9         & 0.3        & 14.6      & 1.9         \\
x-seg         & 0.2        & 25.0      & 1.9         & 0.3        & 20.8      & 2.0         \\
\bottomrule
\end{tabular}
\end{table}

\subsection{Discussion of Findings}
\begin{figure}[ht]
\centering
\includegraphics[width=0.98\linewidth]{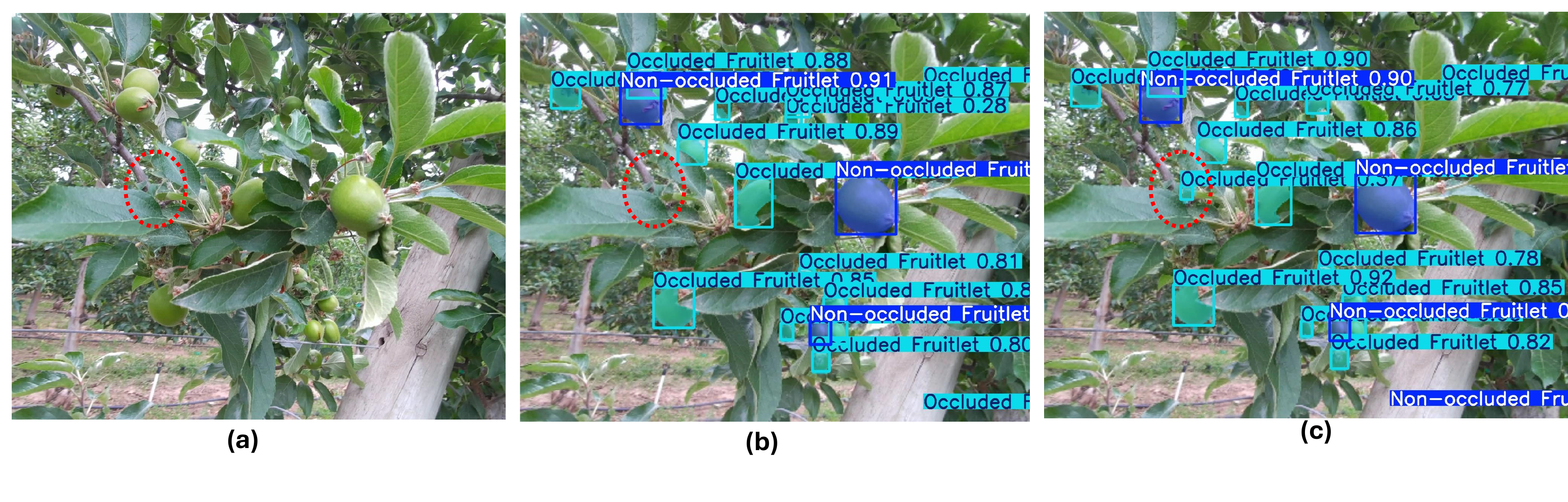}
\caption{Illustrating an example comparison of instance segmentation by YOLO11n-seg and YOLOv8n-seg; (a) shows original RGB images, (b) YOLO11n-seg correctly ignores canopy foliage, whereas (c) YOLOv8n-seg mistakenly identifies it as immature green fruitlets.}
\label{fig:Figure8}
\end{figure}
An observation from the instance segmentation results highlights the precision capabilities of YOLO11 compared to YOLOv8. Figure \ref{fig:Figure8} provides an illustrative comparison, where subfigure \ref{fig:Figure8} (a) displays the original RGB images from a commercial orchard. Subfigures \ref{fig:Figure8} (b) and (c) depict the instance segmentation results for YOLO11n-seg and YOLOv8n-seg, respectively. Notably, a region within these images, marked by a red dotted circle, showcases a significant difference in segmentation accuracy between the two models. This region, characterized by canopy foliage, is accurately disregarded by YOLO11n-seg, which does not erroneously segment this area as a fruit. In contrast, YOLOv8n-seg incorrectly identifies this foliage as an immature green fruitlet, a clear instance of a false positive. 

This discrepancy in model performance can be attributed to several technical enhancements inherent to YOLO11. Firstly, YOLO11 incorporates more sophisticated feature extraction capabilities that may better discriminate between textural and color features of the target objects and their backgrounds. Such advancements could stem from refined training methodologies, deeper network architectures, or more effective learning rate schedules that enhance the model’s ability to generalize from complex backgrounds typical in orchard environments. 

Moreover, YOLO11’s architecture might employ more advanced techniques in handling class imbalance and background noise, crucial for differentiating subtle features in densely packed scenes. The model's ability to ignore non-target regions such as canopy foliage while accurately segmenting true fruitlets indicates superior contextual understanding, which is essential for reducing false positives, a common challenge in agricultural applications. These observations suggest that YOLO11’s design optimizations not only improve its precision but also enhance its practical utility in real-world agricultural settings, where distinguishing between crops and their surrounding environment is paramount. Such capabilities mark a significant step forward in deploying deep learning models for precision agriculture, promising more reliable and efficient automation in fruit detection and counting tasks.

\section{Conclusion and Future Works}
In this study, the latest iterations of the YOLO series, YOLO11 and YOLOv8, along with their configurations: YOLO11n-seg, YOLO11s-seg, YOLO11m-seg, YOLO11l-seg, YOLO11x-seg, YOLOv8n-seg, YOLOv8s-seg, YOLOv8m-seg, YOLOv8l-seg, and YOLOv8x-seg were evaluated for their capability to perform instance segmentation on immature green fruits in commercial orchard environments. All YOLO11 and YOLOv8 models performances were analyzed using precision, recall, and mAP@50 metrics, focusing on segmenting occluded and non-occluded fruitlets. The analysis covered computational requirements, including parameters, GFLOPs, and training layers, providing insights into each model's efficiency and complexity. Additionally, training durations were recorded to gauge time efficiency, vital for real-world agricultural applications. This research offers valuable guidance on selecting the optimal model configuration for precise and rapid detection and segmentation tasks in agricultural settings. From the comprehensive experiment, the specific conclusions of this study are summarized into four following points:

\begin{itemize}
    \item \textbf{Instance Segmentation Metrics for Immature Green Fruits:} YOLO11m-seg showcased the highest mask segmentation performance with mAP@50 scores of 0.860 for "All" and 0.909 for non-occluded fruitlets.
    \item \textbf{Box Detection Metrics for Immature Green Fruits:} In terms of box metrics, YOLO11m-seg achieved the highest mAP@50 scores of 0.876 for "All" class and 0.908 for non-occluded fruitlets.
    \item \textbf{Evaluation of Parameters, GFLOPs, and Layers:} The YOLO11x-seg utilized the most resources, requiring 62.00 million parameters and 318.5 GFLOPs, tailored for high-accuracy segmentation of occluded and non-occluded fruitlets.
    \item \textbf{Training Duration and Image Processing Speeds:} YOLOv8l-seg recorded the longest training time at 5.103 hours, while YOLOv8n exhibited the fastest image processing speed at 3.3 milliseconds, beneficial for real-time applications.
\end{itemize}

The evolution of AI, particularly in the domain of deep learning models like YOLO, necessitates ongoing development to adapt to the rapidly changing technological landscape. Future research should focus on emerging versions (such as future versions which could be YOLOv12, YOLOv13, YOLOv14, YOLOv15, and so on) to assess their performance enhancements in complex agricultural environments. Current models like YOLOv8n demonstrate superior inference speeds compared to YOLO11, especially in scenarios involving occluded immature green fruits in orchards. This suggests a potential for newer models to improve not only in speed but also in detection accuracy and computational efficiency. The continuous iteration of YOLO models could introduce advanced capabilities such as improved noise resistance, better generalization to different fruit types, and enhanced robustness against varying lighting conditions. 

To optimize performance in such an immature green fruit scenario which is a complex environment, future iterations of YOLO should integrate state-of-the-art techniques in machine learning, including advanced neural network architectures and novel training methodologies. For instance, future YOLO versions (future versions may be YOLOv12, YOLOv13, YOLOv14, YOLOv15, and so on) could leverage techniques such as neural architecture search (NAS) \cite{ren2021comprehensive, liu2018progressive} to automatically identify optimal model structures for specific tasks. Moreover, the integration of transfer learning  \cite{weiss2016survey, mellor2021neural, pan2009survey} could expedite training processes and enhance model adaptability across diverse agricultural datasets. Additionally, advancements in unsupervised and semi-supervised learning paradigms could enable these models to perform well even with limited labeled data, which is often a major constraint in agricultural applications. Such developments will be pivotal in pushing the boundaries of what can be achieved with AI in the agricultural sector, leading to more precise, efficient, and cost-effective solutions for real-time fruit detection and segmentation.

Future work will compare YOLO26 or YOLOv26\cite{sapkota2025yolo26, jocher2026ultralytics} and YOLO27 or YOLOv27 \cite{sapkota2025ultralytics, Sapkota2026YOLO27, sapkotaoptimizing} using the same dataset and training protocol.

\section{Acknowledgment and Funding}
This research is funded by the National Science Foundation and United States Department of Agriculture, National Institute of Food and Agriculture through the “AI Institute for Agriculture” Program (Award No.AWD003473). We extend our heartfelt gratitude to Zhichao Meng, Martin Churuvija, Astrid Wimmer, Randall Cason, Diego Lopez, Giulio Diracca, and Priyanka Upadhyaya for their invaluable efforts in data preparation and logistical support throughout this project. Special thanks to Dave Allan for granting orchard access. We also acknowledge the contribution of open-source platforms Roboflow (https://roboflow.com/), Ultralytics (https://docs.ultralytics.com/models/yolo11/ and https://docs.ultralytics.com/), Hugging Face (https://huggingface.co/), and OpenAI (ChatGPT) for the models and implementation assistance in our project through their open-source platform.
\section{Author contributions statement}
R.S data curation, software, methodology, validation,  writing original draft. M.K editing and overall funding to supervisory

\bibliographystyle{unsrtnat}
\bibliography{references}  






\end{document}